\documentclass[final]{cvpr}

\usepackage{times}
\usepackage{epsfig}
\usepackage{graphicx}
\usepackage{amsmath}
\usepackage{amssymb}
\usepackage{mathtools}
\usepackage{enumitem}
\usepackage{xcolor}
\usepackage{xspace}
\usepackage{multirow}

\newcommand{\ve}{\textsf{\small{v2e}}\xspace}

\usepackage[pagebackref=true,breaklinks=true,colorlinks,bookmarks=false]{hyperref}

\usepackage[printwatermark]{xwatermark}
\newwatermark*[page=1,color=gray!60,fontfamily=cmr,fontseries=m,scale=0.4,xpos=0,ypos=120]{This paper has been accepted for publication at the \\ IEEE/CVF Conference on Computer Vision and Pattern Recognition Workshops (CVPRW), Virtual, 2021}

\def\cvprPaperID{5} 
\def\confYear{CVPR 2021}
\begin{document}

\title{v2e: From Video Frames to Realistic DVS Events}

\author{
Yuhuang Hu \qquad Shih-Chii Liu \qquad Tobi Delbruck \\
Institute of Neuroinformatics, University of Z\"urich and ETH Z\"urich, Switzerland \\
{\tt\small\{yuhuang.hu, shih, tobi\}@ini.uzh.ch}
}

\maketitle

\begin{abstract}
To help meet the increasing need for dynamic vision sensor (DVS) event camera data, this paper proposes the \ve toolbox that generates realistic synthetic DVS events from intensity frames. It also clarifies incorrect claims about DVS motion blur and latency characteristics in recent literature. Unlike other toolboxes, \ve includes pixel-level Gaussian event threshold mismatch, finite intensity-dependent bandwidth, and intensity-dependent noise. Realistic DVS events are useful in training networks for uncontrolled lighting conditions. The use of \ve synthetic events is demonstrated in two experiments. The first experiment is object recognition with N-Caltech 101 dataset. Results show that pretraining on various \ve lighting conditions improves generalization when transferred on real DVS data for a ResNet model. The second experiment shows that for night driving, a car detector trained with \ve events shows an average accuracy improvement of 40\% compared to the YOLOv3 trained on intensity frames.
\end{abstract}

\section{Introduction}
\label{sec:intro}

A Dynamic Vision Sensor (\textbf{DVS}) outputs brightness change events~\cite{lichtsteiner2008128,Gallego2019-cm}. Each pixel holds a memorized brightness value (log intensity value) and continuously monitors if the brightness changes away from this stored value by a specified event threshold.
The high dynamic range, fine time resolution, and quick, sparse output make DVS attractive sensors for machine vision under difficult lighting conditions and limited computing power. 
Since the first DVS cameras, subsequent generations of DVS-type event cameras have been developed; see~\cite{delbruck2010activity, Liu2014-oy, Posch2014-ieee-proc, Gallego2019-cm} for surveys.

With the growing commercial development of event cameras and the application of deep learning to the camera output, large DVS datasets are needed for training these networks. Although the number of DVS datasets is growing (see~\cite{github-event-based-vision-resources}), they are still far fewer than frame-camera datasets. Thus, DVS simulators\cite{katz2012live, Mueggler2017-event-cam-dataset, rebecq-esim-corl18} and transfer learning methods such as~\cite{rebecq-esim-corl18,rpg:video:to:events:Gehrig:2020} were developed to exploit existing intensity and mixed modality~\cite{nga:Hu:2020} frame datasets.

Computer vision papers about event cameras have made incorrect claims such as ``event cameras [have] no motion blur'' and have ``latency on the order of microseconds''~\cite{Scheerlinck2020-fastnet,rebecq-esim-corl18,Mitrokhin2019-hh}, perhaps fueled by the titles of papers like~\cite{lichtsteiner2008128,Brandli2014-cs,Serrano-Gotarredona2013-3p6us-latency}, which report their best metrics obtained under lab conditions. Recent reviews like~\cite{Gallego2019-cm} are not explicit about the actual behavior under low light conditions. DVS cameras must obey the laws of physics like any other vision sensor: Their output is based on counting photons. Under low illumination conditions, photons become scarce, and therefore counting them becomes noisy and slow.

This paper introduces the \ve toolbox that is aimed at realistic modeling of these conditions and crucial for the deployment of event cameras in uncontrolled lighting conditions.
The main contributions of this work are as follows:

\begin{figure*}[ht]
\centering
\includegraphics[width=0.8 \linewidth]{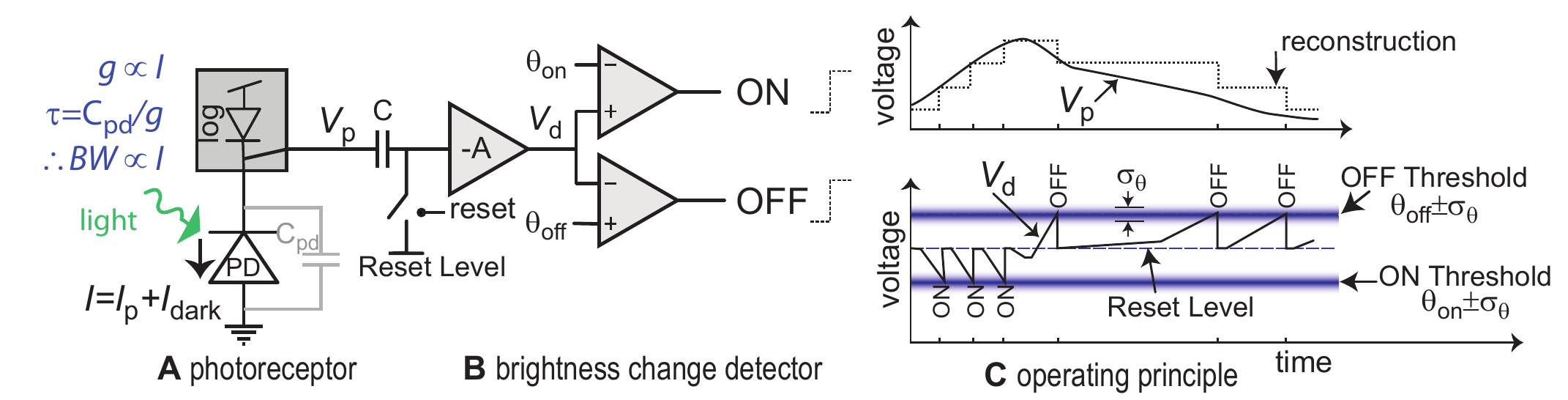}
   \caption{Simplified DVS pixel with principle of operation. Light falling on the photodiode (PD) creates a photocurrent, $I_p$. More details in text. BW=pixel bandwidth, $\theta_{\rm ON}$=ON threshold, $\theta_{\rm OFF}$=OFF thresholds, $V_p$=photoreceptor output voltage, $V_d$= amplified voltage output. Figure adapted from~\cite{lichtsteiner2008128,Nozaki2017-bt}.}
\label{fig:dvsprinciple}
\end{figure*}

\begin{enumerate}
    \item A description of the operation of the DVS pixel for the computer vision community, together with the behavior of DVS pixels under low illumination (Sec.~\ref{sec:dvspixel});
    \item A demystification of claims in the computer vision literature about lack of motion blur and DVS latency (Secs.~\ref{sec:dvspixel:motion:blur}, \ref{sec:dvspixel:latency});
    \item The \ve toolbox\footnote{Link: \url{https://github.com/SensorsINI/v2e}}, which introduces the first DVS pixel model that includes temporal noise, leak events, finite intensity-dependent bandwidth, and Gaussian threshold distribution (Sec.~\ref{sec:method});
    \item A newly labeled dataset MVSEC-NIGHTL21\footnote{Link: \url{http://sensors.ini.uzh.ch/databases.html}} from a subset of the MVSEC dataset for car detection in night driving conditions (Sec.~\ref{sec:method:datasets}).
    \item Sec.~\ref{sec:results} shows that a network benefits from training with low-light \ve synthetic events. Network generalization is improved when transferring to real event data.
\end{enumerate}

In addition to realistic \ve visual examples in Sec.~\ref{sec:results:demo}, we use \ve toolbox in two computer vision tasks. We first study object recognition in Sec.~\ref{sec:results:ncaltech} using N-Caltech 101 dataset \cite{ncaltech:Orchard:2015}. The result shows that by training on \ve events synthesized in multiple lighting conditions, our classification network surpasses the supervised baseline accuracy after fine-tuning. The second task is a car detection task using the MVSEC dataset~\cite{mvsec:Zhu:2018}. By combining the Network Grafting Algorithm (\textbf{NGA})~\cite{nga:Hu:2020} and synthetic events, we trained a detection network that works well in the day condition. In the night condition, this detector exhibits up to 40\% accuracy improvement compared to YOLOv3, which is an intensity frame-based detector (Sec.~\ref{sec:results:mvsec}).

\section{Prior Work for DVS Simulators}
Katz et al.~\cite{katz2012live} used a 200\,Hz frame rate camera to synthesize DVS events with a 5\,ms time resolution using a simple model of the DVS pixel operation that generated DVS events from the camera intensity samples. The \textit{Event Camera Dataset and Simulator}~\cite{Mueggler2017-event-cam-dataset} and the newer \textit{ESIM}~\cite{rebecq-esim-corl18} toolboxes can be used to generate synthetic DVS events from synthetic video (\eg, using Blender) or image datasets, and thus enabled many recent advances in processing DVS output based on transfer learning. An extension to \textit{ESIM} called \textit{rpg\_vid2e} drives \textit{ESIM} from interpolated video frames~\cite{rpg:video:to:events:Gehrig:2020}. \textit{rpg\_vid2e} used the same idealistic model of DVS pixels as \textit{ESIM}, that is, it assumed that the DVS pixel bandwidth is at least as large as the upsampled video rate, there is no temporal noise, no leak DVS events, and that the threshold mismatch is uniformly distributed --- all of which are invalid for real cameras. Thus \textit{rpg\_vid2e} simulated ideal DVS pixels under good lighting, but not realistic DVS pixels under bad lighting, which is an important use case for DVS. The \ve toolbox proposed in this paper is a step towards incorporating a more realistic DVS model in the simulator. By enabling explicit control of the noise and nonideality `knobs', \ve enables the generation of synthetic datasets covering a range of illumination conditions.

\section{DVS Pixel Operation and Biases} 
\label{sec:dvspixel}

Fig.~\ref{fig:dvsprinciple} shows a simplified schematic of the DVS pixel circuit. The continuous-time process of generating events is illustrated in Fig.~\ref{fig:dvsprinciple}C. The DVS pixel bias current parameters control the pixel event threshold and analog bandwidth. In Fig.~\ref{fig:dvsprinciple}A, the input photocurrent generates a continuous logarithmic photoreceptor output voltage $V_{\rm p}$. The change amplifier in Fig.~\ref{fig:dvsprinciple}B produces an inverted and amplified output voltage $V_{\rm d}$. When  $V_{\rm d}$ crosses either the ON or OFF threshold voltage, the pixel emits an event (via a shared digital output that is not shown). The event reset memorizes the new log intensity value across the capacitor $C$.

The logarithmic response of the photoreceptor comes from the exponential current versus voltage relationship in the feedback diode (gray box in Fig.~\ref{fig:dvsprinciple}A). The smaller the photocurrent $I$, the longer the time constant $\tau=C_\text{pd}/g$, where $C_\text{pd}$ is the photodiode capacitance and $g\propto I$ is the conductance of the feedback diode. The event thresholds $\theta_\text{ON}$ and $\theta_\text{OFF}$ are nominally identical across pixels but statistically vary by $\sigma_{\theta}$ because of transistor mismatch.

\subsection{DVS under low lighting}
\begin{figure}[ht]
\centering
\includegraphics[width=0.85\linewidth]{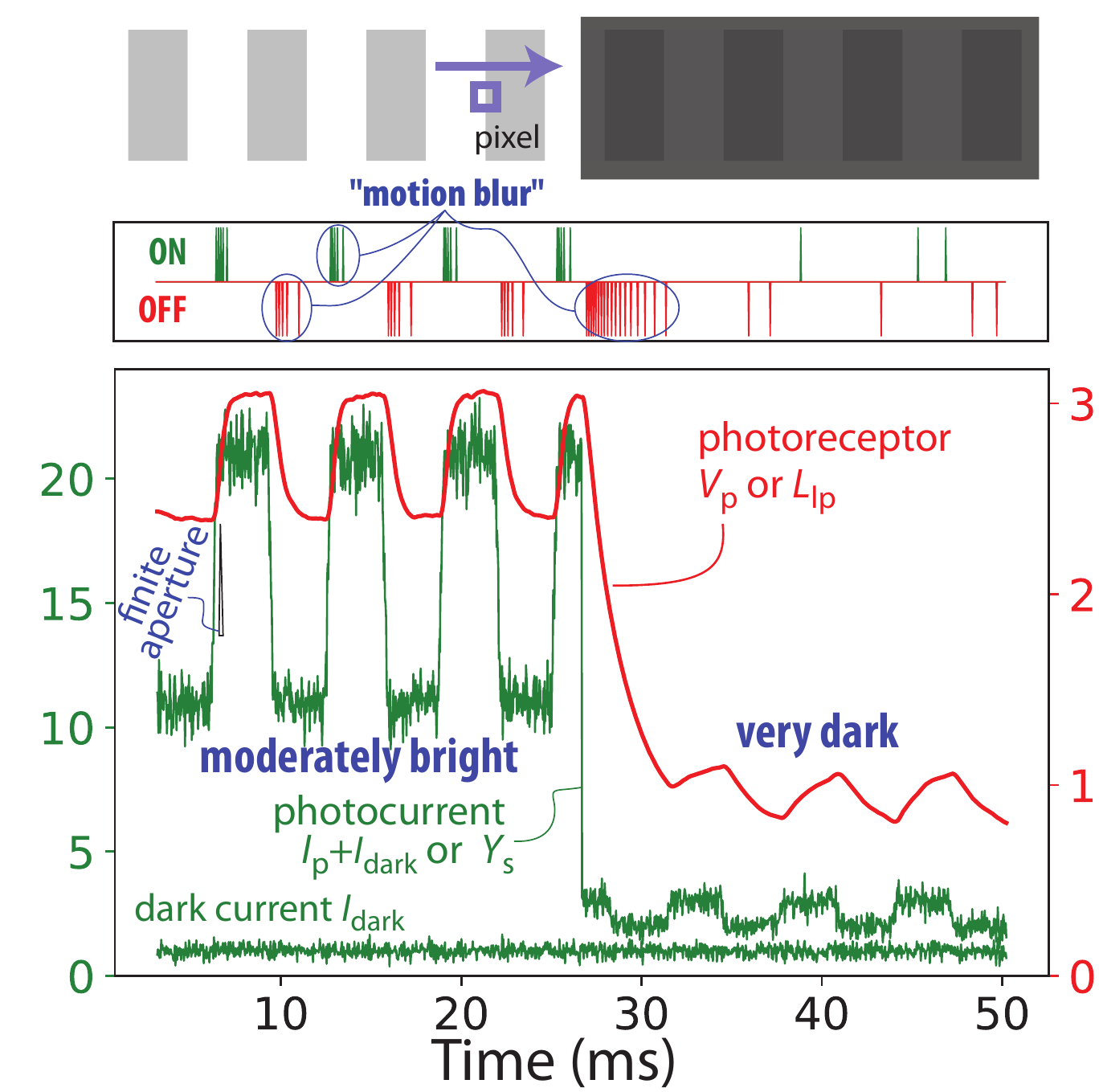}
   \caption{Simulated DVS pixel photoreceptor and resulting ON and OFF events under moderate and extremely low illumination. Both photocurrent and dark current include shot noise which is proportional to the mean current.\\
   Code to reproduce: \url{https://git.io/JOWbG}}
\label{fig:dvs-low-lighting}
\end{figure}

Fig.~\ref{fig:dvs-low-lighting} shows a behavioral simulation based on logarithmic photoreceptor dynamics of a DVS pixel operating under extremely low illumination conditions when a grating consisting of alternating gray and white strips passes over the pixel. Note that in the absence of light, a continuous dark current $I_\text{dark}$ flows through the photodiode. During the initial ``moderately bright'' cycles, the signal photocurrent, $I_p\gg I_\text{dark}$ and the bandwidth of the photoreceptor, which depends on $I_p$, is high enough so that  $V_p$  can follow the input current fluctuations. The contrast of the signal was set to 2 so that the white part of the grating produced twice the photocurrent compared to the gray part. The pixel makes about five events for each rising and falling edge (the change threshold was set to 0.1 units), but these are spread over time due to the rise and fall time of $V_p$. In the shadowed ``very dark'' section, the overall illumination is reduced by 10X. The contrast of the \emph{signal} is still unchanged (the reflectance of the scene is the same as before), but now $I_p$ is comparable to $I_\text{dark}$, thus reducing the actual contrast of the current fluctuations. Because $I$ is so small, the bandwidth decreases to the point where the photoreceptor can no longer follow the input current fluctuations and the edges become extremely motion blurred. Both effects reduce the number of generated brightness change events (to about 2 per edge) and increase their timing jitter. \ve models these effects to produce realistic low-light synthetic DVS events.

\subsection{Motion blur}\label{sec:dvspixel:motion:blur}
For frame-based video, motion blur is simply a lowpass box filter imposed by the finite integration time for the frame.
It should be obvious from Fig.~\ref{fig:dvs-low-lighting} that a DVS pixel does not respond instantly to an edge: The finite response time of the photoreceptor blurs the edge. The transition from one brightness level to another is like the response of an RC lowpass filter. The bigger the step, the longer it takes for the pixel to settle to the new brightness value. The result is that a passing edge will result in an extended series of events as the pixel settles down to the new value. 
This finite response time over which the pixel continues to emit events is the equivalent ``motion blur'' of DVS pixels. Under bright indoor illumination, typical values for the pixel motion blur are on the order of 1\,ms. Under very low illumination, the equivalent pixel motion blur can extend for tens of milliseconds.

\begin{figure}[ht]
\centering
\includegraphics[width=\linewidth]{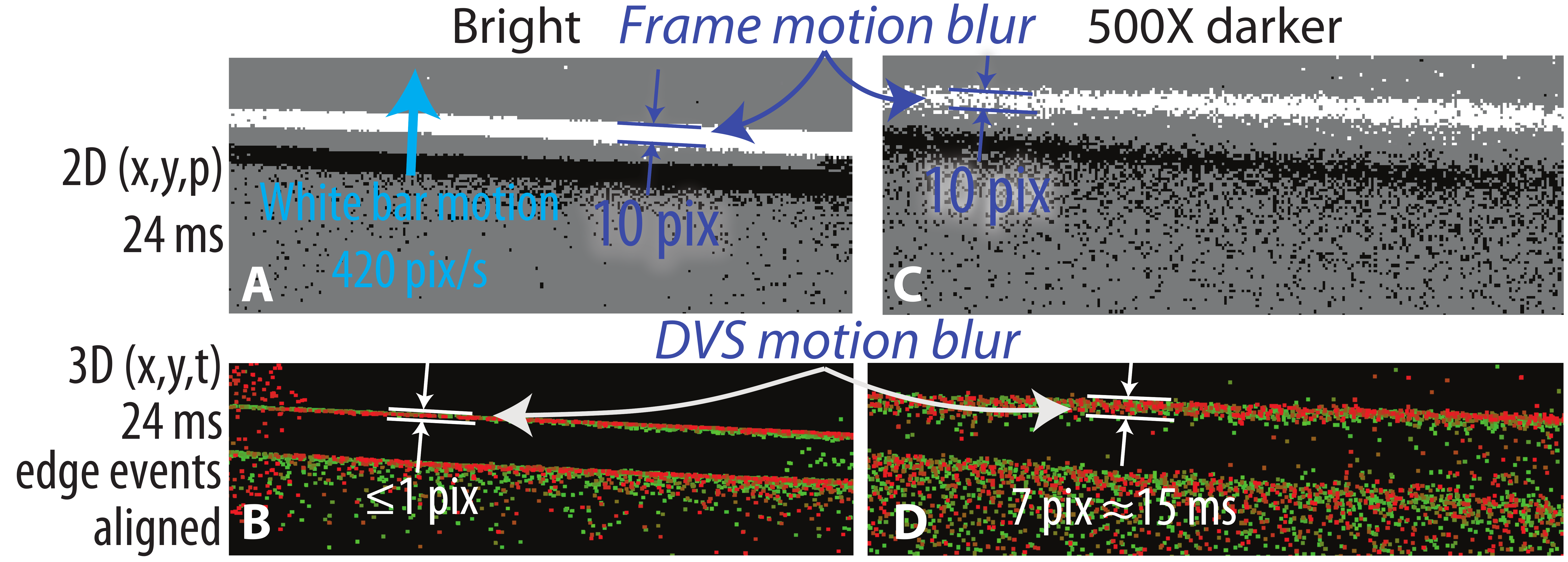}
   \caption{Measured motion blur of real DVS outputs for a moving white bar (speed: 420 pixels/s) on a dark background.}
\label{fig:dvs-motion-blur}
\end{figure}

Fig.~\ref{fig:dvs-motion-blur} shows measured DAVIS346~\cite{Taverni2018-fsi-bsi-davis346} DVS motion blur of a moving edge under bright and dark conditions. 
Users typically view DVS output as 2D frames of histogrammed event counts collected over a fixed integration time (Fig.~\ref{fig:dvs-motion-blur}A and C). The frame integration time low-pass filters the DVS output stream just like conventional video cameras. The additional DVS motion blur can be easily observed by lining up the events in a 3D space-time view of the event cloud that compensates for the motion of the edge (Figs.~\ref{fig:dvs-motion-blur}B and D). In this view, the DVS motion blur appears as a thickened edge. In Fig.~\ref{fig:dvs-motion-blur}B, the motion blur of the leading white edge is less than 1 pixel (\ie, less than 2\,ms), but in Fig.~\ref{fig:dvs-motion-blur}D, the blur is about 7 pixels or 15\,ms.

\subsection{Latency}\label{sec:dvspixel:latency}
Quick response time is a clear advantage of DVS cameras, and they have been used to build complete visually servoed robots with total closed-loop latencies of under 3\,ms~\cite{Delbruck2013-robogoalie,Conradt2009-pencil}. But it is important to realize the true range of achievable response latency. For example, high-speed USB computer interfaces impose a minimum latency of a few hundred microseconds~\cite{Delbruck2013-robogoalie}.

\begin{figure}[ht]
    \centering
    \includegraphics[width= 0.8\columnwidth]{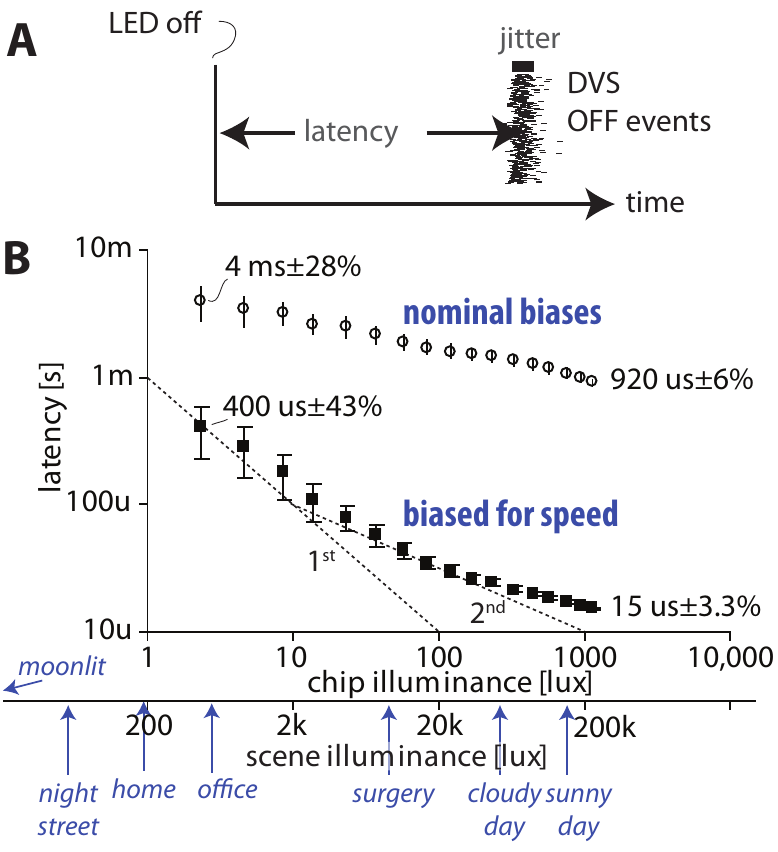}
   \caption{Real DVS latency measurements to turning off blinking LED. \textbf{A:}~definition of response latency. \textbf{B:}~measured data. Adapted from~\cite{lichtsteiner2008128}, with scene illumination axis based on~\cite{Delbruck1997-photometry}.}
\label{fig:DVS128_JSSC_2008_latency+jitter}
\end{figure}

\begin{figure*}[ht]
\begin{center}
\includegraphics[width=.8\linewidth]{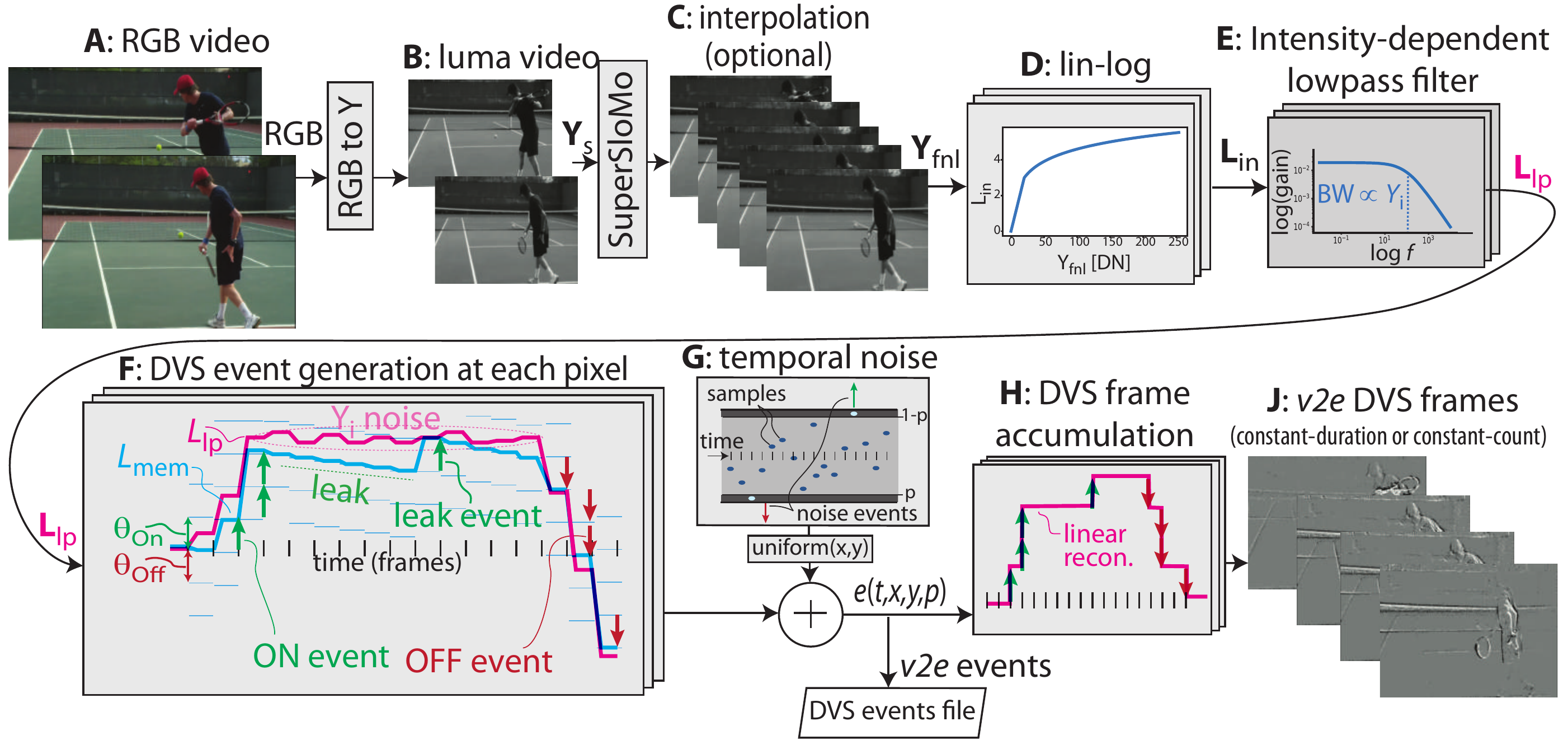}
\end{center}
   \caption{Steps of the \ve DVS event generation (Sec.~\ref{sec:method}; see lettered headers).}
\label{fig:alg_steps}
\end{figure*}

Added to these computer and operating system latencies are the DVS sensor chip latencies, which are illustrated in Fig.~\ref{fig:DVS128_JSSC_2008_latency+jitter} with real DVS data. This experiment recorded the response latency to a blinking LED turning off. The horizontal axis in Fig.~\ref{fig:DVS128_JSSC_2008_latency+jitter}B is in units of lux (visible photons/area/time): The upper scale is for chip illumination, and the lower scale is for scene illumination, assuming 20\% scene reflectance and a \textit{f}/2.8 lens aperture ratio~\cite{Delbruck1997-photometry}. Typical scenarios are listed below the scene illuminance axis. The DVS was biased in two different ways: The ``nominal biases'' setup used settings that are meant for everyday use of the DVS. With these settings, the DVS pixel bandwidth is limited by the photoreceptor and source follower biases, and thus the DVS latency is only a soft function of intensity. This choice limits noise at low light intensities. The ``biased for speed'' setup uses higher bias currents for the photoreceptor and source follower to optimize the DVS for the quickest possible response, with the tradeoff of additional noise from a shorter integration time. With this setup, we see that the latency decreases with the reciprocal of intensity. Typical users of DVS will experience real-world latencies in the order of about one to a few ms and latency jitter in the order of 100\,$\mu$s to 1\,ms. The absolute minimum latency is reported in papers as a figure of merit for such sensors (as is customary in the electronics community), but it clearly does not reflect real-world use. Additional discussion on DVS latency can be found in~\cite{inivation:report:2020}.
The \ve lowpass filtering (Sec~\ref{sec:method}) models these effects.

\section{The \ve Toolbox}
\label{sec:method}
Fig.~\ref{fig:alg_steps} shows the steps of the DVS emulation starting from RGB pixel intensity samples.

\textbf{A-B: Color to luma conversion:}
\ve starts from a source video that is $T$ seconds long. The frames of color video are automatically converted into $M$ luma frames, $\mathbf{Y}_{\rm s}=\{\mathbf{Y}_{\rm s}^{(i)}\}_{i=1}^{M}$, using the ITU-R recommendation \textit{BT.\,709} digital video (linear, non-gamma-corrected) color space conversion~\cite{itu:luma:2015}. Each frame $\mathbf{Y}_{\rm s}^{(i)}$ is associated with a timestamp $t_{i}$ where $0=t_{1}<\ldots<t_{i}<\ldots<t_{M}=T$. For grayscale video frames, the pixel value is treated as the luma value. After conversion to luma, frames are optionally scaled to the desired output height and width in pixels.

\textbf{C: Synthetic slow motion:}
The luma frames are then optionally interpolated using the \textit{Super-SloMo} video interpolation network~\cite{jiang2018super} to increase the temporal resolution of the input video. Super-SlowMo predicts the bi-directional optic flow vectors from consecutive luma frames that are then used to linearly interpolate new frames at arbitrary times between the two input frames. To better estimate flow for luma frames, we retrained Super-SloMo on the Adobe240FPS~\cite{su2017deep} dataset after converting its RGB frames to luma frames.

If upsampling is not needed, we define the upsampling ratio $U_{\rm fnl}=1$. Alternatively, to determine $U_{\rm fnl}$, the user can choose both a maximum DVS timestamp step $\Delta t_{\max}$ and whether to activate automatic upsampling that was introduced in~\cite{rpg:video:to:events:Gehrig:2020}. The manual upsampling ratio $U_{\rm man}$ is computed from the source video frame rate $f_{\rm s}$, $U=\lceil1/(f_{\rm s} \times \Delta t_{\max})\rceil$. For example, $f_{\rm s}=60\rm Hz$ and $\Delta t_{\max}=1\rm ms$ result in $U_{\rm man}=17$. If automatic upsampling is activated, \ve computes the maximum optic flow $F$ in pixels over a batch of frames from the Super-SloMo optic flow estimate. It then computes $U_{\rm fnl}=\max(U_{\rm man},\lceil F\rceil)$ to limit the maximum flow per interframe to 1~pixel. The frame rate after the optional upsampling is $f_{\rm fnl}=f_{s}\times U_{\rm fnl}$. The upsampled frames $\mathbf{Y}_{\rm fnl}=\{\mathbf{Y}_{\rm fnl}^{(j)}\}_{j=1}^{M\times U_{\rm fnl}}$ corresponds to timestamps $0=t_{1}<\ldots<t_{j}<\ldots<t_{M\times U_{\rm fnl}}=T$.

For simplicity, the following discusses synthetic event generation for a single DVS pixel. We denote $Y$ as the pixel's luma intensity value in a luma frame $\mathbf{Y}$. Similarly, we use $L$ to represent the pixel's log intensity values in a log intensity frame $\mathbf{L}$.

\textbf{D: Linear to logarithmic mapping:} The method to generate events from frames is based on~\cite{katz2012live}. Standard digital video usually represents intensity linearly, but DVS pixels detect changes in log intensity. By default, computer vision uses 8-bit values, equivalent to  a limited dynamic range (DR) of $255=48\text{dB}$. To deal with this limited DR and quantization, we use a lin-log mapping between $Y$ and log intensity value $L$ as illustrated in Fig.~\ref{fig:alg_steps}D. For luma intensity value $Y<20$ digital number (\textbf{DN}), we use a linear mapping from exposure value (intensity) to log intensity. The linearizing part of the conversion function means that small $Y$ values will be converted linearly, reducing quantization noise in the synthetic DVS output.

\textbf{E: Finite intensity-dependent photoreceptor bandwidth:} Since the real DVS pixel has finite analog bandwidth, an optional lowpass filter filters the input $L$ value. This filter models the DVS pixel response under low illumination as discussed in Sec.~\ref{sec:dvspixel}. The DVS pixel bandwidth is proportional to intensity, at least for low photocurrents~\cite{lichtsteiner2008128}. \ve models this effect for each pixel by making the filter bandwidth (BW) increase monotonically with the intensity value. Although the photoreceptor and source follower form a 2nd-order lowpass, one pole is usually dominant, and so this filter is implemented by an infinite impulse response (IIR) first-order lowpass filter. The nominal cutoff frequency is $f_{\rm 3dBmax}$ for full white pixels. The filter's bandwidth is proportional to the luma intensity values $Y$. We denote the filtered $L$ value $L_{\rm lp}$. The shape of the filter's transfer function is illustrated in Fig.~\ref{fig:alg_steps}E.

To avoid nearly zero bandwidth for small DN pixels, an additive constant limits the minimum bandwidth to about 10\% of the maximum value. The update is done by the steps in the supplementary material, along with details of the filter.

\textit{\textbf{Logarithm and temporal contrast threshold:}} We define the pixel event thresholds for generating ON and OFF events as $\theta_{\rm ON}>0$, $\theta_{\rm OFF}<0$. Typically the magnitudes of $\theta_{\rm ON}$ and $\theta_{\rm OFF}$ are quite similar and take on values from $0.1<|\theta|<0.4$, \ie, the typical range of adjustable DVS thresholds is approximately from 10\% to 50\% light intensity change. That is, the change of the logarithmic value $\Delta L=L_{\rm new}-L_{\rm old}=\ln(Y_{\rm new}/Y_{\rm old})$ corresponds to the intensity change ratio, $Y_{\rm new}/Y_{\rm old}$.

The event thresholds are dimensionless and represent a threshold for relative intensity change, \ie, a threshold on the change of the intensity by a ratio relative to the memorized value. These relative intensity changes are produced by scene reflectance changes, which is why this representation is useful for producing events that are informative about the visual input.

\textbf{F: Event generation model:}
We assume that the pixel has a memorized brightness value $L_{\rm mem}$ in log intensity and that the new low pass filtered brightness value is $L_{\rm lp}$. The model then generates a signed integer quantity $N_{\rm e}$ of positive ON or negative OFF events from the change $\Delta L=L_{\rm lp}-L_{\rm mem}$ where $N_{\rm e}=\left\lfloor\frac{\Delta L}{\theta}\right\rfloor$. Details are in the supplementary material.

If $\Delta L$ is a multiple of the ON and OFF thresholds, multiple DVS events are generated. The memorized brightness value is updated by $N_{\rm e}$ multiples of the threshold.

\textit{\textbf{Threshold mismatch:}} The typical value of DVS contrast threshold is about $|\theta_{\rm nominal}|=0.3$. Measurements show that the threshold varies with a Gaussian~\cite{lichtsteiner2008128} distribution  $\sigma_{\theta}\approx 3\%$ contrast, \ie, before starting the DVS event generation, we store a 2D array of $\theta_{\rm ON}$ and $\theta_{\rm OFF}$ values drawn from $\theta_{\rm nominal}+\mathcal{N}(0,\sigma_{\theta})$ with $\sigma_{\theta}=0.03$.

\textit{\textbf{Hot pixels:}} DVS sensors always have some `hot pixels', which continuously fire events at a high rate even in the absence of input. Hot pixels can result from abnormally low thresholds or reset switches with a very high dark current. Hot pixels are created by the frozen threshold sampling, but \ve limits the minimum threshold to 0.01 to prevent too many hot pixel events.

\textit{\textbf{Leak  noise events:}} DVS pixels emit spontaneous ON events called \textit{leak events}~\cite{Nozaki2017-bt} with typical rates $\approx$ 0.1\,Hz. They are caused by junction leakage and parasitic photocurrent in the change detector reset switch~\cite{Nozaki2017-bt}. \ve adds these leak events by continuously decreasing the memorized brightness value $L_{\rm mem}$ as shown in Fig.~\ref{fig:alg_steps}F. The leak rate varies according to random variations of the event threshold, which decorrelates leak events from different pixels.

\textbf{G: Temporal noise:}
The quantal nature of photons results in \emph{shot noise}: If, on average, $K$ photons are accumulated in each integration period, then the average variance will also be $K$. At low light intensities, the effect of this shot noise on DVS output events increases dramatically, resulting in balanced ON and OFF shot noise events at above 1\,Hz per pixel rate. \ve models temporal noise using a Poisson process. It generates ON and OFF temporal noise events to match a noise event rate $R_n$ (default 1\,Hz). To model the increase of temporal noise with reduced intensity, the noise rate $R_{n}$ is multiplied by a linear function of luma $0<Y\leq1$ that reduces noise in bright parts by a factor $0<c<1$ (default $c=0.25$). This modified rate $r$ is multiplied by the time step $\Delta t$ to obtain the probability $p=r\times\Delta t\ll1$ that will be applied to the next sample. For each sample, a uniformly distributed number in the range 0-1 is compared against two thresholds $[p,1-p]$ as illustrated in Fig.~\ref{fig:alg_steps}G
to decide if an ON or OFF noise event is generated. These noise events are added to the output and reset the pixels. The complete steps are described in the supplementary material.

\textit{\textbf{Event timestamps:}} The timestamps of the interpolated frames are discrete. Given two consecutive interpolated frames, the timestamps of events are evenly distributed in between $(t_{j}, t_{j+1})$.

\section{Data Preparation and Datasets}
\subsection{Event voxel grid representation}
\label{sec:method:ev:voxel}

Our experiments use the event voxel grid method to convert $N$ events into a 3D representation with size $H\times W\times D$~\cite{etovid:Rebecq:2019, nga:Hu:2020} to use as the network input for the Sec.~\ref{sec:results} results. $H$ and $W$ are sensor height and width dimensions. $D$ is a hyperparameter that defines the number of slices of the output voxel grid. These slices are effectively frames of histogrammed DVS events where each slice has an exposure time of $(N/R)/D$, where $R$ is the average event rate.

\subsection{Datasets}\label{sec:method:datasets}

\textbf{N-Caltech 101~\cite{ncaltech:Orchard:2015}} is an event-based object recognition dataset generated from the Caltech 101 object recognition dataset~\cite{caltech101:Li:2004}. Each image in the Caltech 101 dataset was recorded by a DVS for 300\,ms using three 100\,ms-long triangular saccades. The dataset contains 8,709 object samples over 101 object categories. For experiments in Sec.~\ref{sec:results:ncaltech}, we synthesized \ve synthetic datasets under three different conditions, namely, Ideal (no non-ideality), Bright (modest amount of noise), and Dark (most noise). The synthesis parameters are in Table.~\ref{tab:v2e:params}. Each recording of the synthetic dataset has the same duration and number of saccades as in the original N-Caltech 101 dataset. When preparing event voxel grids, we choose $D=15$ slices and use all events of each recording. Each slice in an event voxel grid corresponds to 20\,ms. (If the DVS motion blur is smaller than the slice duration, it would not be easily noticeable.)

\textbf{MVSEC~\cite{mvsec:Zhu:2018}} is a stereo driving dataset captured from two DAVIS346 event cameras. The cameras recorded both intensity frames and events. We used the intensity frames in \texttt{outdoor\_day\_2} to synthesize the events for the day condition (parameters in Table~\ref{tab:v2e:params}). We set $N=25,000$ events and $D=10$ when preparing event voxel grids. This setting is identical to \cite{nga:Hu:2020}. There are 5,900 intensity frame and event voxel grid pairs generated from the \ve day recording. These pairs are used as training samples for NGA. For experiments in Sec.~\ref{sec:results:mvsec}, we also used the training and validation datasets from ~\cite{nga:Hu:2020} that were generated from the real events. Additionally, we generate 2,000 pairs from the MVSEC \texttt{outdoor\_night\_1} recording. The first 1,600 pairs are used as the night training samples. The night recording in MVSEC is not labeled. We hand-labeled the remaining 400 pairs for cars to create a night validation set.

\begin{table}[ht]
    \centering
    \caption{\ve synthesis parameters for N-Caltech 101 and MVSEC datasets. $\theta$ and $\sigma_{\theta}$ represent event threshold and threshold variation respectively.}\label{tab:v2e:params}
    \resizebox{\linewidth}{!}{
    \begin{tabular}{c|ccc|c}
        \hline
        \textbf{Dataset} & \multicolumn{3}{|c|}{\textbf{N-Caltech 101}}  & \multicolumn{1}{|c}{\textbf{MVSEC}}\\
        \hline
        \textbf{Condition} & Ideal & Bright & Dark &  Day\\
        \hline
        \multirow{2}{*}{$\theta$} & \multirow{2}{*}{0.05-0.5} & \multirow{2}{*}{0.05-0.5} & \multirow{2}{*}{0.05-0.5} & $(0.73_{\text{ON}},$ \\
        & & & & $0.43_{\text{OFF}})$ \\
        $\sigma_{\theta}$ & $\times$\textsuperscript{\dag} & $\mathcal{U}(\cdot,\cdot)$\textsuperscript{\ddag} & $\mathcal{U}(\cdot,\cdot)$ & 0.03\\
        \textbf{Shot Noise} & $\times$ & $\times$ & 1-10 Hz & 2\,Hz\\
        \textbf{Leak Events} & $\times$ & 0.1-0.5\,Hz & $\times$ & 0.5\,Hz\\
        \textbf{Cutoff Freq.} & $\times$ & 200\,Hz & 10-100\,Hz & 200\,Hz \\
        \hline
        \multicolumn{5}{l}{\dag\,\,`$\times$' sign means not applicable.}\\
        \multicolumn{5}{l}{\ddag\,\,$\mathcal{U}(\cdot,\cdot)=\text{Uniform} (\min(0.15\,\theta, 0.03),\min(0.25\,\theta,0.05))$.} \\
        \hline
    \end{tabular}
    }
\end{table}

\section{Results}
\label{sec:results}
\subsection{Qualitative demonstrations}
\label{sec:results:demo}

\begin{figure}[ht]
\centering
\includegraphics[width=0.8\linewidth]{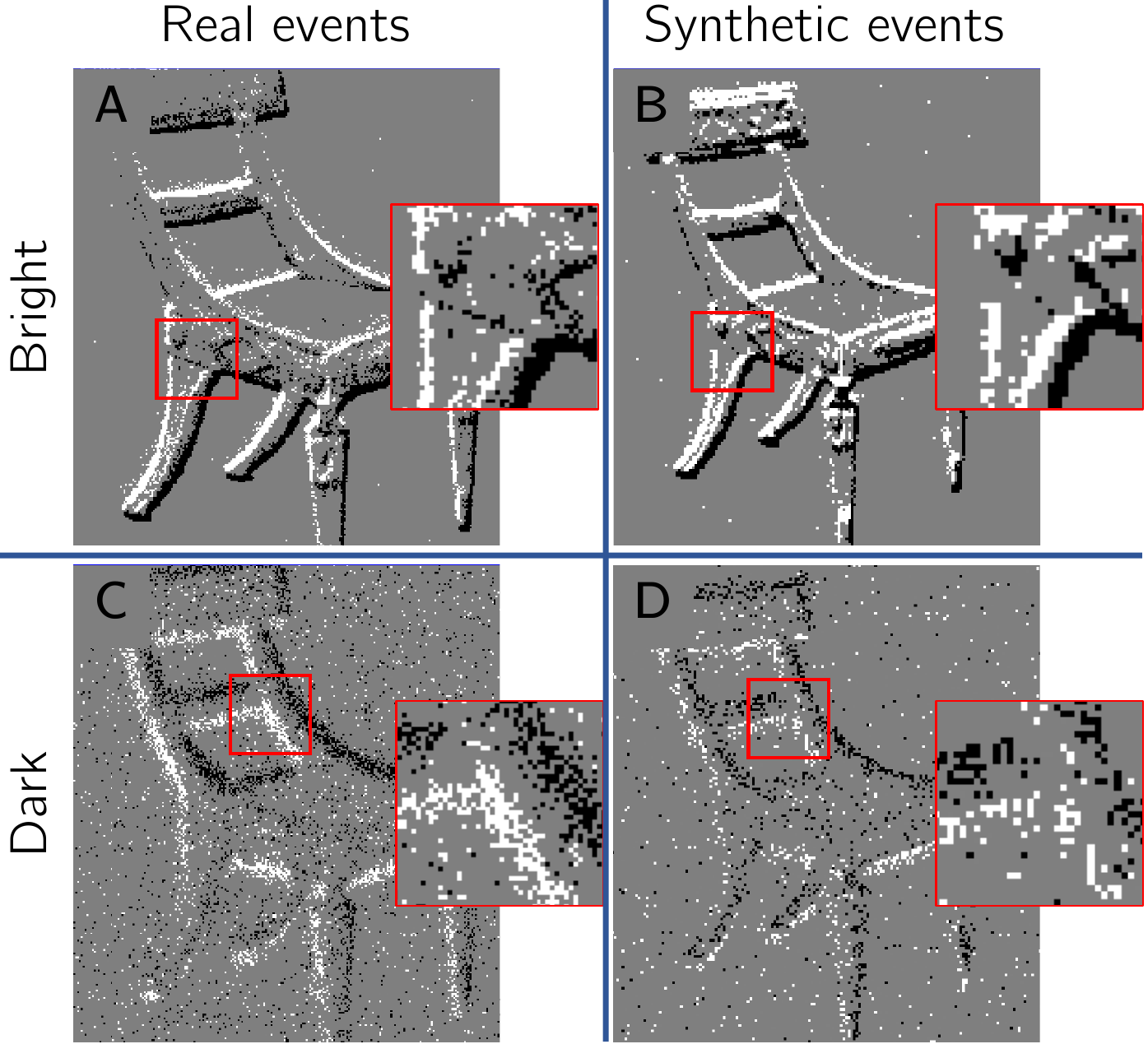}

   \caption{\texttt{chair} example under bright and dark conditions. \textbf{A} and \textbf{C} are generated from real events; \textbf{B} and \textbf{D} are from \ve synthetic events. Zoom-in views of the selected regions are also displayed.}
\label{fig:chair-comparison}
\end{figure}

We use a \texttt{chair} example from the N-Caltech 101 dataset to demonstrate qualitatively that \ve can realistically synthesize DVS events under bright and dark illumination. Fig.~\ref{fig:chair-comparison} compares 20\,ms snapshots of the DVS output. We displayed the 300\,ms chair video at a high frame rate on a monitor (after ensuring the monitor had no backlight flicker) and recorded the DAVIS346 DVS output with the lens aperture fully open (Fig.~\ref{fig:chair-comparison}A) and closed down to 450X lower luminance (Fig.~\ref{fig:chair-comparison}C). Then we modeled DVS events from the chair video with \ve using bright (Fig.~\ref{fig:chair-comparison}B) and dark settings (Fig.~\ref{fig:chair-comparison}D). The motion blurring and noise under low illumination are clearly visible and \ve produces qualitatively similar effects as reduced lighting.

\subsection{N-Caltech 101 object recognition} \label{sec:results:ncaltech}

This section presents experiments that show the use of \ve synthetic data in an N-Caltech 101 object recognition task. Unless otherwise mentioned, models in this section were trained for 100 epochs with a batch size of 4 using AdamW~\cite{adamw:Loshchilov:2019} optimizer. The initial learning rate is $10^{-4}$, and the learning rate decreases by 10 at every 30 epochs. The recognition network is a ResNet34~\cite{resnet:He:2016} pretrained on ImageNet. We replaced the first layer to accommodate the voxel grid input and re-initialized the classification layer. For every class of each dataset, we randomly selected 45\% as the train samples, 30\% as validation samples, and 25\% as test samples. This data split is fixed for all datasets in this section. Every model was evaluated on the same test dataset that contains real event recordings from the original N-Caltech 101 dataset.

\begin{figure}[ht]
    \centering
    \includegraphics[width=\linewidth]{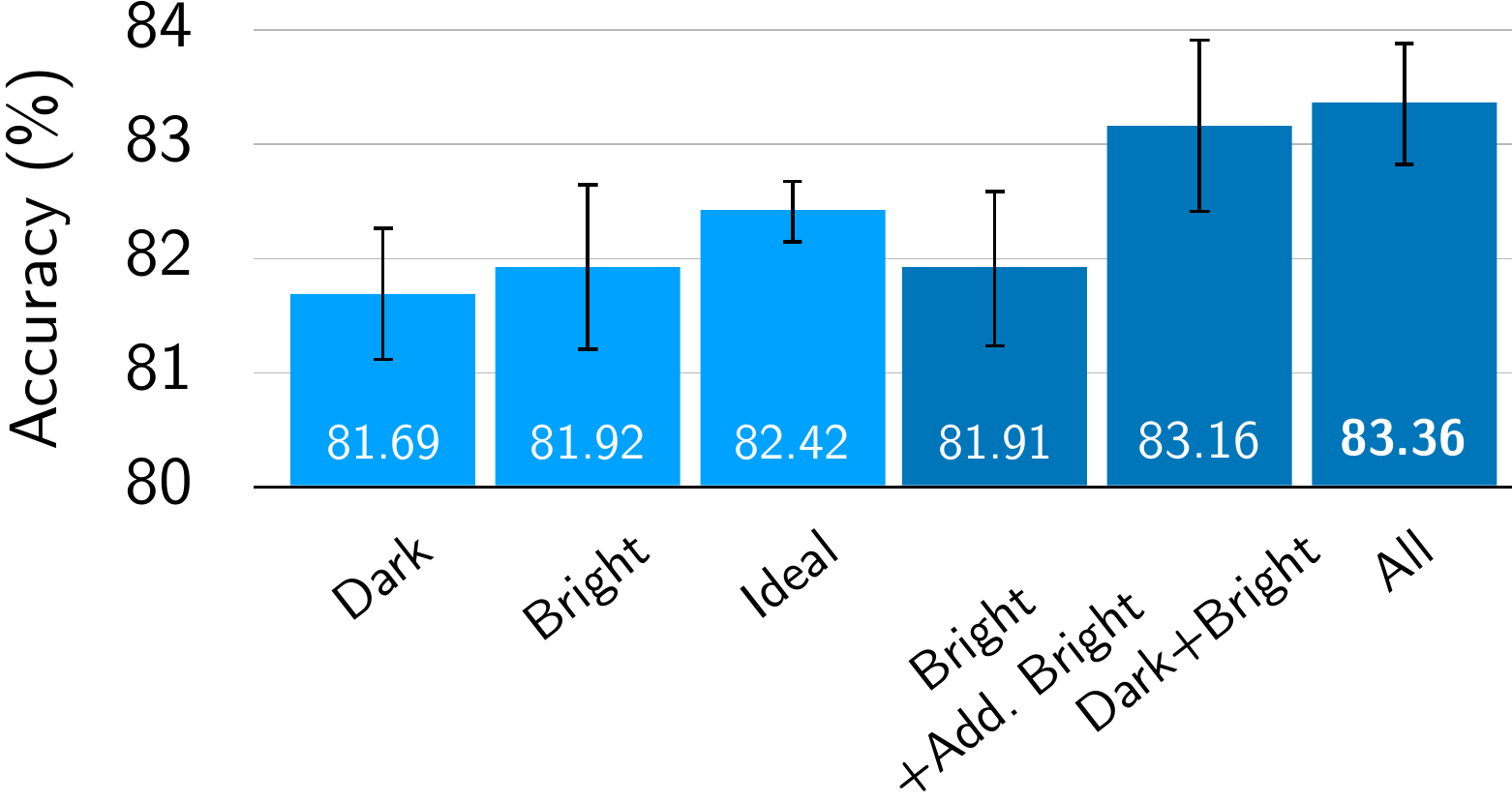}
    \caption{N-Caltech 101 test accuracy. The \texttt{x}-axis indicates different \ve synthetic training datasets and combinations. The five-run averaged accuracy is showed at the bottom of each bar.}
    \label{fig:ncaltech:v2e:result}
\end{figure}

Fig.~\ref{fig:ncaltech:v2e:result} studies the effect of the event camera non-idealities. With training data from events synthesized under the three conditions: Ideal, Bright, and Dark, we found that the accuracy of the ResNet34 network is highest for the Ideal condition because there is no noise. Because the original N-Caltech 101 was recorded in bright conditions, this result was expected. However, by combining synthesized events from two (Bright+Dark) or three conditions for the training data, the test accuracy is better than when training on a single condition. For instance, 83\% for All \vs 82\% for Ideal. To make the accuracy scores in Fig.~\ref{fig:ncaltech:v2e:result} comparable, the number of epochs for the combined datasets were reduced proportionally so that the same total number of samples were presented during training. We also synthesized additional \ve samples under Bright condition (labeled as Add.~Bright), which are added to the previously generated Bright dataset to form a new `Bright' training set. From Fig.~\ref{fig:ncaltech:v2e:result}, it is clear that the accuracies are similar (Bright \vs Bright+Add. Bright). This result shows that by including a wide range of synthesis parameters, the gap between network accuracy from real and synthetic event data can be reduced.

\begin{table}[ht]
    \centering
    \caption{N-Caltech 101 test accuracy on the classification task. Reported results are averaged over five runs.}\label{tab:ncaltech:results}
    \begin{tabular}{lc|c}
        \hline
        \multicolumn{1}{c}{\textbf{Method}} & \textbf{Training Dataset} & \textbf{Accuracy (\%)}  \\
        \hline
        ResNet34 (ours) & real events & 86.74$\pm$0.54 \\
        ResNet34 (ours) & \ve-All & 83.36$\pm$0.76\\
        ResNet34 (ours) & +fine-tune & \textbf{87.85$\pm$0.12} \\
        \hline
        HATS~\cite{hats:ncaltech:Sironi:2018} & real events & 64.20\\
        RG-CNNs~\cite{rgcnn:ncaltech:Bi:2019} & real events & 65.70 \\
        EST~\cite{est:ncaltech:Gehrig:2019} & real events & 81.70\\
        \hline
        ResNet34 \cite{rpg:video:to:events:Gehrig:2020} & real events & 86.30 \\
        ResNet34 \cite{rpg:video:to:events:Gehrig:2020} & synthetic events & 78.20 \\
        ResNet34 \cite{rpg:video:to:events:Gehrig:2020} & +fine-tune & \textbf{90.40} \\
         \hline
    \end{tabular}
\end{table}

\begin{figure*}[ht]
    \centering
    \includegraphics[width=0.8\linewidth]{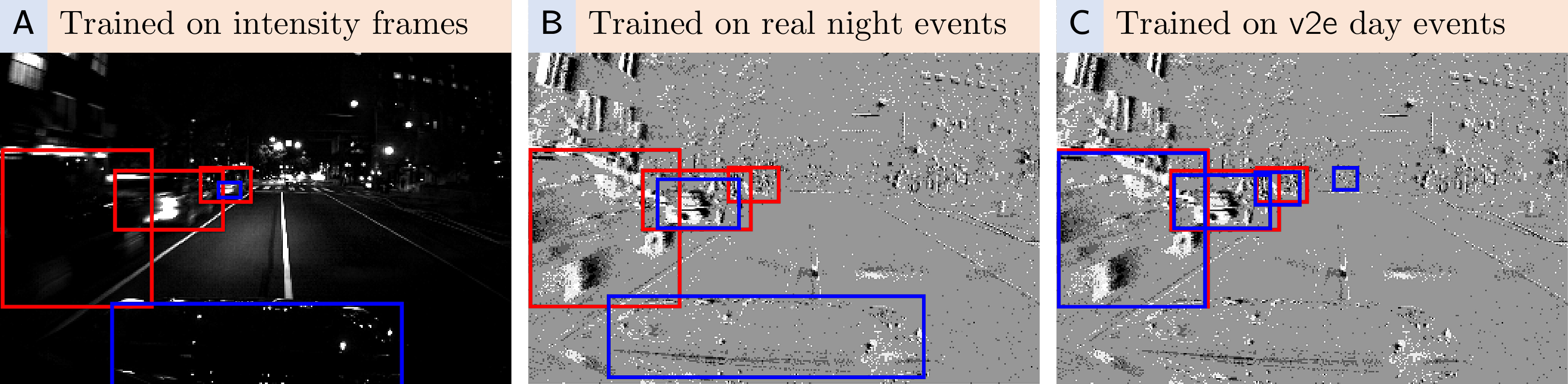}
    \caption{Car detection examples on the MVSEC night recording. The groundtruth bounding boxes are in red while the predicted boxes in blue. The detection quality in \textbf{C} is visually better than \textbf{B} and \textbf{A}.}
    \label{fig:mvsec:night:compare}
\end{figure*}

Table~\ref{tab:ncaltech:results} presents three groups of accuracies on the N-Caltech 101 dataset. In the first group, we first established the baseline (86.74\%) by training a model using real events. By combining three synthesis conditions, the model trained only on synthetic data reaches 83.36\%, which is only 3\% lower than the baseline. After training with synthetic data, we fine-tuned this trained model on the real events until convergence with learning rate $10^{-6}$. This fine-tuned model reaches 87.85\% accuracy, which is significantly better than the baseline. This group of results shows that the pretraining on various \ve conditions improves model generalization on real event data. Compared to the second group of the table, our fine-tuned model accuracy is also higher than recent literature~\cite{hats:ncaltech:Sironi:2018, rgcnn:ncaltech:Bi:2019, est:ncaltech:Gehrig:2019}. The most similar prior art is Ref.~\cite{rpg:video:to:events:Gehrig:2020} and their results are summarized in the third group of Table.~\ref{tab:ncaltech:results}. With the same ResNet34 model, our baseline and synthetic data trained model accuracies are higher than theirs. However, our model accuracy after fine-tuning is lower than their model that reached 90.4\%. (The Ref.~\cite{rpg:video:to:events:Gehrig:2020}'s dataset split and training code are not available, so we could not replicate their results.)

\subsection{MVSEC car detection}\label{sec:results:mvsec}

In this section, we used \ve synthetic data to train a car detection network using the Network Grafting Algorithm (\textbf{NGA})~\cite{nga:Hu:2020}. The NGA algorithm enables training of an event-driven network using paired synchronous intensity frames and brightness change events (here generated by \ve). We followed the same setup and the training schedule as in~\cite{nga:Hu:2020} using the YOLOv3 detection network~\cite{yolo:Redmon:2018}.

\begin{table}[ht]
    \centering
    \caption{MVSEC car detection results in average precision. The five-run averaged scores are shown. \texttt{GN} refers to grafted network based on YOLOv3.}\label{tab:mvsec:results}
    \begin{tabular}{ll|c}
        \hline
        \multicolumn{1}{c}{\textbf{Model}} & \multicolumn{1}{c}{\textbf{Training Dataset}} & \multicolumn{1}{|c}{AP$_{50}$} \\
        \hline
        \multicolumn{3}{c}{\textbf{Test on real day events}} \\
        \hline
        \texttt{GN-B1}~\cite{nga:Hu:2020} & real day events & 70.35$\pm$0.51 \\
        \texttt{GN-D1} (ours) & \ve day events & 62.52$\pm$1.15 \\
        \texttt{GN-D2} (ours) & +fine-tune & 69.82$\pm$0.64 \\
        \hline
        \texttt{GN-B2}~\cite{nga:Hu:2020} & real day events (10\%) & 47.88$\pm$1.86 \\
        \texttt{GN-D3} (ours) & +fine-tune (10\%) & 68.55$\pm$0.28 \\
        \hline
        \multicolumn{3}{c}{\textbf{Test on real night intensity frames}} \\
        \hline
        YOLOv3 & intensity frames & 25.94 \\
        \hline
        \multicolumn{3}{c}{\textbf{Test on real night events}} \\
        \hline
        \texttt{GN-B1}~\cite{nga:Hu:2020} & real day events & 35.67 \\
        \texttt{GN-N1} (ours) & real night events & 29.38$\pm$1.08 \\
        \texttt{GN-D1} (ours) & \ve day events & \textbf{36.41$\pm$2.90} \\
        \hline
    \end{tabular}
\end{table}

Table~\ref{tab:mvsec:results} summarizes our findings in two groups. In the first group, compared to the model trained with the real day events, our model \texttt{GN-D1} trained with \ve day events gives an average precision ($\text{AP}_{50}$) that is 11\% lower than baseline \texttt{GN-B1}'s accuracy 70.35. We fine-tuned \texttt{GN-D1} with the real day events for 10 epochs to reduce this accuracy gap. We find that the fine-tuned model \texttt{GN-D2}'s accuracy of 69.82 is on par with the baseline \texttt{GN-B1}~\cite{nga:Hu:2020}. We also fine-tuned \texttt{GN-D1} with only 10\% of real day event data (labelled as \texttt{GN-D3}). Compared to~\cite{nga:Hu:2020}, \texttt{GN-D3}'s accuracy of 68.55 is 43\% higher than the baseline \texttt{GN-B2}'s $\text{AP}_{50}$ of 47.88. This result shows that pretraining on \ve synthetic events is beneficial for model generalization.

The second study (results in Table~\ref{tab:mvsec:results}) is to determine whether the DVS allows the network output to be more invariant to different illumination levels compared to intensity cameras. A recent paper~\cite{event:object:detection:Etienne:2020} only observed this phenomenon qualitatively. Here, we used our newly labeled real night validation set (see Sec.~\ref{sec:method:datasets}) to quantitatively validate this hypothesis. First, we evaluated the accuracy of the original YOLOv3 on the night intensity frames for the baseline $\text{AP}_{50}$ of 25.94. Second, by NGA training on paired real night intensity frames and events, the grafted network \texttt{GN-N1} reaches an improved accuracy of 29.38, showing that the higher DR of the event camera is useful. Finally, we used model \texttt{GN-D1}. The $\text{AP}_{50}$ is 36.41, which is 40\% better than 25.94 that is obtained by using the original YOLOv3. The \texttt{GN-D1}'s accuracy is also higher than a baseline accuracy which is obtained by running the \texttt{GN-B1} model.

\begin{figure}[ht]
    \centering
    \includegraphics[width=\linewidth]{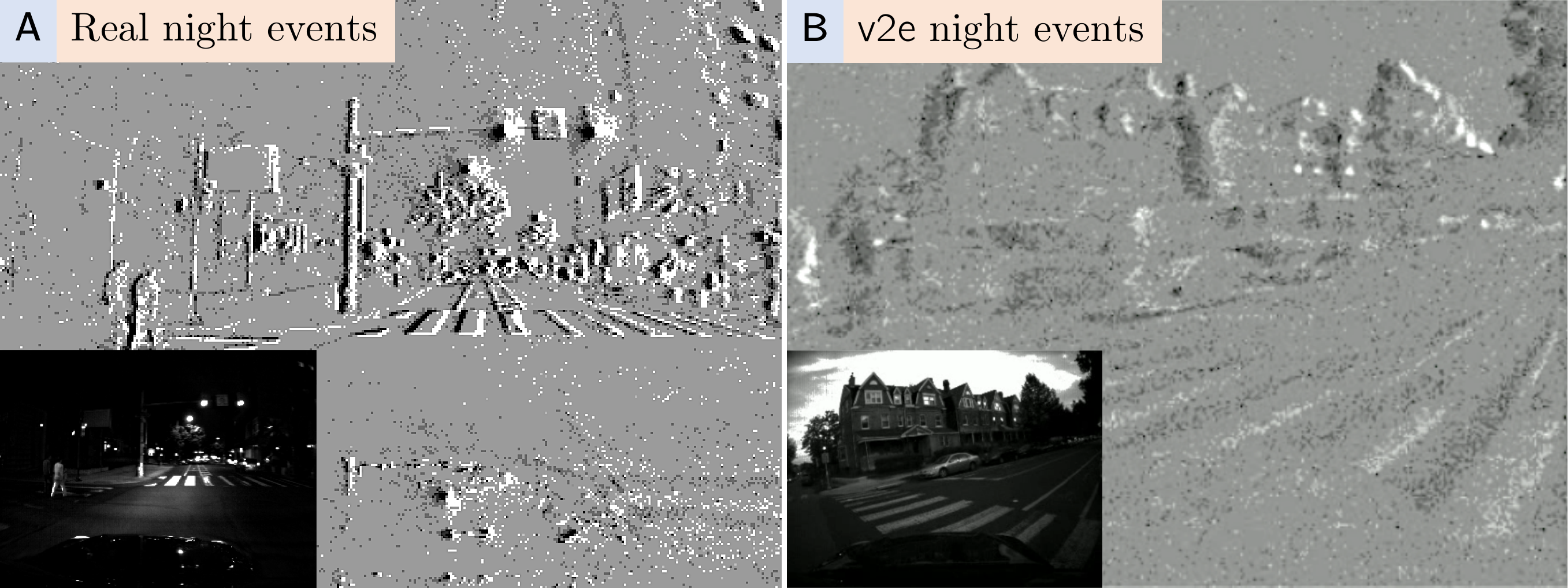}
    \caption{\textbf{A}: the headlight of the car illuminates mostly the center of the scene; \textbf{B}: \ve motion-blurred the entire scene uniformly.}
    \label{fig:mvsec:compare:motion:blur}
\end{figure}

The improved result of \texttt{GN-D1} for night condition can be explained as follows: \textbf{1.}~NGA works better for high-quality intensity frames and sharp DVS event inputs, but the real night intensity frames are severely underexposed and motion blurred (Fig.~\ref{fig:mvsec:night:compare}A, B); \textbf{2.}~because the event camera is more robust to different lighting conditions than intensity cameras, \texttt{GN-D1} trained with better exposed \ve day intensity frames transferred better to the night scenes (Fig.~\ref{fig:mvsec:night:compare}C).

We also used \ve to synthesize night events from the MVSEC day recording by setting the cutoff frequency to 10\,Hz. We were surprised to find that by training with these synthetic night events, the accuracy is even lower than the $\text{AP}_{50}$ of the original YOLOv3. This counterintuitive result can be understood with Fig.~\ref{fig:mvsec:compare:motion:blur}. In the night recording, the front area was illuminated by the car's headlight (Fig.~\ref{fig:mvsec:compare:motion:blur}A) while \ve uniformly motion-blurred the entire scene (Fig.~\ref{fig:mvsec:compare:motion:blur}B). The grafted network trained on the synthetic night events could not infer sharp features as available in the real night events.

\section{Conclusion} \label{sec:discuss}

This paper described the \ve toolbox that can synthesize realistic events from intensity frames. By modeling the noise and motion blur non-idealities of event cameras, we hope to debunk common myths about the event camera and bridge the gap between the simulation model and the real sensor. With this tool, we hope to stimulate more research in understanding and modeling these non-idealities.

Our experiments showed qualitative and quantitative evidence that \ve can generate realistic DVS events and that the synthetic events are useful. Along with simulators such as \textit{ESIM}~\cite{rebecq-esim-corl18} and its extension~\cite{rpg:video:to:events:Gehrig:2020}, \ve provides practical ways to sample a large amount of diverse synthetic DVS events for building more robust event-driven algorithms. 

Compared to \cite{event:object:detection:Etienne:2020} that only showed visual examples, this paper, for the first, time proved that the event camera is more robust under low lighting conditions for a car detection task. The results are encouraging for further research in using event cameras under difficult lighting conditions.

\paragraph{Acknowledgments} This work was funded by the Swiss National Competence
Center in Robotics (NCCR Robotics).

{\small
\bibliographystyle{ieee_fullname}
\bibliography{egbib}

\begin{thebibliography}{10}\itemsep=-1pt

\bibitem{rgcnn:ncaltech:Bi:2019}
Y. {Bi}, A. {Chadha}, A. {Abbas}, E. {Bourtsoulatze}, and Y. {Andreopoulos}.
\newblock Graph-based object classification for neuromorphic vision sensing.
\newblock In {\em 2019 IEEE/CVF International Conference on Computer Vision
  (ICCV)}, pages 491--501, 2019.

\bibitem{Brandli2014-cs}
C. {Brandli}, R. {Berner}, M. {Yang}, S. {Liu}, and T. {Delbruck}.
\newblock A 240$\times$180 130 db 3 $\mu$s latency global shutter
  spatiotemporal vision sensor.
\newblock {\em IEEE Journal of Solid-State Circuits}, 49(10):2333--2341, 2014.

\bibitem{Conradt2009-pencil}
J. {Conradt}, M. {Cook}, R. {Berner}, P. {Lichtsteiner}, R.~J. {Douglas}, and
  T. {Delbruck}.
\newblock A pencil balancing robot using a pair of aer dynamic vision sensors.
\newblock In {\em 2009 IEEE International Symposium on Circuits and Systems},
  pages 781--784, 2009.

\bibitem{Delbruck2013-robogoalie}
T. Delbruck and M. Lang.
\newblock Robotic goalie with 3 ms reaction time at 4\% {CPU} load using
  event-based dynamic vision sensor.
\newblock {\em Frontiers in Neuroscience}, 7:223, 2013.

\bibitem{Delbruck1997-photometry}
T. Delbruck and N. Mascarenhas.
\newblock Notes on practical photometry.
\newblock \url{https://www.ini.uzh.ch/~tobi/wiki/doku.php?id=radiometry},
  1997--2017.

\bibitem{delbruck2010activity}
T. {Delbrück}, B. {Linares-Barranco}, E. {Culurciello}, and C. {Posch}.
\newblock Activity-driven, event-based vision sensors.
\newblock In {\em Proceedings of 2010 IEEE International Symposium on Circuits
  and Systems}, pages 2426--2429, 2010.

\bibitem{Gallego2019-cm}
G. {Gallego}, T. {Delbruck}, G.~M. {Orchard}, C. {Bartolozzi}, B. {Taba}, A.
  {Censi}, S. {Leutenegger}, A. {Davison}, J. {Conradt}, K. {Daniilidis}, and
  D. {Scaramuzza}.
\newblock {Event-based Vision}: {A} survey.
\newblock {\em IEEE Transactions on Pattern Analysis and Machine Intelligence},
  pages 1--1, 2020.

\bibitem{rpg:video:to:events:Gehrig:2020}
D. {Gehrig}, M. {Gehrig}, J. {Hidalgo-Carrió}, and D. {Scaramuzza}.
\newblock {Video to Events: R}ecycling video datasets for event cameras.
\newblock In {\em 2020 IEEE/CVF Conference on Computer Vision and Pattern
  Recognition (CVPR)}, pages 3583--3592, 2020.

\bibitem{est:ncaltech:Gehrig:2019}
D. {Gehrig}, A. {Loquercio}, K. {Derpanis}, and D. {Scaramuzza}.
\newblock End-to-end learning of representations for asynchronous event-based
  data.
\newblock In {\em 2019 IEEE/CVF International Conference on Computer Vision
  (ICCV)}, pages 5632--5642, 2019.

\bibitem{resnet:He:2016}
K. {He}, X. {Zhang}, S. {Ren}, and J. {Sun}.
\newblock Deep residual learning for image recognition.
\newblock In {\em 2016 IEEE Conference on Computer Vision and Pattern
  Recognition (CVPR)}, pages 770--778, 2016.

\bibitem{nga:Hu:2020}
Y. {Hu}, T. Delbruck, and S-C. {Liu}.
\newblock Learning to exploit multiple vision modalities by using grafted
  networks.
\newblock In A. Vedaldi, H. Bischof, T. Brox, and J-M. Frahm, editors, {\em
  Computer Vision -- ECCV 2020}, pages 85--101, Cham, 2020. Springer
  International Publishing.

\bibitem{inivation:report:2020}
{iniVation}.
\newblock Understanding the performance of neuromorphic event-based vision
  sensors.
\newblock Technical report, iniVation, May 2020.

\bibitem{itu:luma:2015}
{International Telecomunication Union}.
\newblock Parameter values for the hdtv standards for production and
  international programme exchange.
\newblock Technical report, International Telecomunication Union, 2015.
\newblock BT.709.

\bibitem{jiang2018super}
H. {Jiang}, D. {Sun}, V. {Jampani}, M. {Yang}, E. {Learned-Miller}, and J.
  {Kautz}.
\newblock {Super SloMo: H}igh quality estimation of multiple intermediate
  frames for video interpolation.
\newblock In {\em 2018 IEEE/CVF Conference on Computer Vision and Pattern
  Recognition}, pages 9000--9008, 2018.

\bibitem{katz2012live}
M.~L. {Katz}, K. {Nikolic}, and T. {Delbruck}.
\newblock Live demonstration: {B}ehavioural emulation of event-based vision
  sensors.
\newblock In {\em 2012 IEEE International Symposium on Circuits and Systems
  (ISCAS)}, pages 736--740, 2012.

\bibitem{caltech101:Li:2004}
{Li Fei-Fei}, R. {Fergus}, and P. {Perona}.
\newblock Learning generative visual models from few training examples: {A}n
  incremental bayesian approach tested on 101 object categories.
\newblock In {\em 2004 Conference on Computer Vision and Pattern Recognition
  Workshop}, pages 178--178, 2004.

\bibitem{lichtsteiner2008128}
P. {Lichtsteiner}, C. {Posch}, and T. {Delbruck}.
\newblock A 128$\times$ 128 120 db 15 $\mu$s latency asynchronous temporal
  contrast vision sensor.
\newblock {\em IEEE Journal of Solid-State Circuits}, 43(2):566--576, 2008.

\bibitem{Liu2014-oy}
S-C. Liu, T. Delbruck, G. Indiveri, A. Whatley, and R. Douglas.
\newblock {\em {Event-Based} Neuromorphic Systems}.
\newblock John Wiley \& Sons, Dec. 2014.

\bibitem{adamw:Loshchilov:2019}
I. Loshchilov and F. Hutter.
\newblock Decoupled weight decay regularization.
\newblock In {\em International Conference on Learning Representations}, 2019.

\bibitem{Mitrokhin2019-hh}
A. {Mitrokhin}, C. {Ye}, C. {Fermüller}, Y. {Aloimonos}, and T. {Delbruck}.
\newblock {EV-IMO: M}otion segmentation dataset and learning pipeline for event
  cameras.
\newblock In {\em 2019 IEEE/RSJ International Conference on Intelligent Robots
  and Systems (IROS)}, pages 6105--6112, 2019.

\bibitem{Mueggler2017-event-cam-dataset}
E. Mueggler, H. Rebecq, G. Gallego, T. Delbruck, and D. Scaramuzza.
\newblock The event-camera dataset and simulator: {E}vent-based data for pose
  estimation, visual odometry, and slam.
\newblock {\em The International Journal of Robotics Research}, 36(2):142--149,
  2017.

\bibitem{Nozaki2017-bt}
Y. {Nozaki} and T. {Delbruck}.
\newblock Temperature and parasitic photocurrent effects in dynamic vision
  sensors.
\newblock {\em IEEE Transactions on Electron Devices}, 64(8):3239--3245, 2017.

\bibitem{ncaltech:Orchard:2015}
G. {Orchard}, A. {Jayawant}, G.~K. {Cohen}, and N. {Thakor}.
\newblock Converting static image datasets to spiking neuromorphic datasets
  using saccades.
\newblock {\em Frontiers in Neuroscience}, 9:437, 2015.

\bibitem{event:object:detection:Etienne:2020}
E. Perot, P. de Tournemire, D. Nitti, J. Masci, and A. Sironi.
\newblock Learning to detect objects with a 1 megapixel event camera.
\newblock In H. Larochelle, M. Ranzato, R. Hadsell, M.~F. Balcan, and H. Lin,
  editors, {\em Advances in Neural Information Processing Systems}, volume~33,
  pages 16639--16652. Curran Associates, Inc., 2020.

\bibitem{Posch2014-ieee-proc}
C. {Posch}, T. {Serrano-Gotarredona}, B. {Linares-Barranco}, and T. {Delbruck}.
\newblock {Retinomorphic Event-Based Vision Sensors: B}ioinspired cameras with
  spiking output.
\newblock {\em Proceedings of the IEEE}, 102(10):1470--1484, 2014.

\bibitem{rebecq-esim-corl18}
H. Rebecq, D. Gehrig, and D. Scaramuzza.
\newblock {ESIM}: an open event camera simulator.
\newblock In Aude Billard, Anca Dragan, Jan Peters, and Jun Morimoto, editors,
  {\em Proceedings of The 2nd Conference on Robot Learning}, volume~87 of {\em
  Proceedings of Machine Learning Research}, pages 969--982. PMLR, 29--31 Oct
  2018.

\bibitem{etovid:Rebecq:2019}
H. {Rebecq}, R. {Ranftl}, V. {Koltun}, and D. {Scaramuzza}.
\newblock {Events-To-Video: B}ringing modern computer vision to event cameras.
\newblock In {\em 2019 IEEE/CVF Conference on Computer Vision and Pattern
  Recognition (CVPR)}, pages 3852--3861, 2019.

\bibitem{yolo:Redmon:2018}
J. Redmon and A. Farhadi.
\newblock {YOLOv3: A}n incremental improvement.
\newblock {\em CoRR}, abs/1804.02767, 2018.

\bibitem{Scheerlinck2020-fastnet}
C. {Scheerlinck}, H. {Rebecq}, D. {Gehrig}, N. {Barnes}, R.~E. {Mahony}, and D.
  {Scaramuzza}.
\newblock Fast image reconstruction with an event camera.
\newblock In {\em 2020 IEEE Winter Conference on Applications of Computer
  Vision (WACV)}, pages 156--163, 2020.

\bibitem{Serrano-Gotarredona2013-3p6us-latency}
T. {Serrano-Gotarredona} and B. {Linares-Barranco}.
\newblock A 128$\,\times$ 128 1.5\% contrast sensitivity 0.9\% fpn 3 $\mu$s
  latency 4 mw asynchronous frame-free dynamic vision sensor using
  transimpedance preamplifiers.
\newblock {\em IEEE Journal of Solid-State Circuits}, 48(3):827--838, 2013.

\bibitem{hats:ncaltech:Sironi:2018}
A. {Sironi}, M. {Brambilla}, N. {Bourdis}, X. {Lagorce}, and R. {Benosman}.
\newblock {HATS: H}istograms of averaged time surfaces for robust event-based
  object classification.
\newblock In {\em 2018 IEEE/CVF Conference on Computer Vision and Pattern
  Recognition}, pages 1731--1740, 2018.

\bibitem{su2017deep}
S. {Su}, M. {Delbracio}, J. {Wang}, G. {Sapiro}, W. {Heidrich}, and O. {Wang}.
\newblock Deep video deblurring for hand-held cameras.
\newblock In {\em 2017 IEEE Conference on Computer Vision and Pattern
  Recognition (CVPR)}, pages 237--246, 2017.

\bibitem{Taverni2018-fsi-bsi-davis346}
G. {Taverni}, D. {Paul Moeys}, C. {Li}, C. {Cavaco}, V. {Motsnyi}, D. {San
  Segundo Bello}, and T. {Delbruck}.
\newblock Front and back illuminated dynamic and active pixel vision sensors
  comparison.
\newblock {\em IEEE Transactions on Circuits and Systems II: Express Briefs},
  65(5):677--681, 2018.

\bibitem{github-event-based-vision-resources}
{Various contributors}.
\newblock Event-based vision resources.
\newblock \url{https://github.com/uzh-rpg/event-based_vision_resources}, 2020.

\bibitem{mvsec:Zhu:2018}
A.~Z. {Zhu}, D. {Thakur}, T. {Özaslan}, B. {Pfrommer}, V. {Kumar}, and K.
  {Daniilidis}.
\newblock {The Multivehicle Stereo Event Camera Dataset: A}n event camera
  dataset for 3d perception.
\newblock {\em IEEE Robotics and Automation Letters}, 3(3):2032--2039, 2018.

\end{thebibliography}
}
\end{document}